\pdfoutput=1
\documentclass[5p]{elsarticle}

\usepackage{amsmath}
\usepackage{amssymb}
\usepackage[dvipsnames]{xcolor}

\usepackage{todonotes}

\usepackage{algorithm}
\usepackage{algpseudocode}
\algblockdefx[Input]{Instart}{Inend}{\textbf{inputs:}}{}
\algblockdefx[Output]{Outstart}{Outend}{\textbf{outputs:}}{}

\usepackage{amsthm}
\newtheorem{definition}{Definition}

\newcommand{\x}{\boldsymbol{x}}
\newcommand{\xroot}{\widetilde{\x}}

\newcommand{\sv}{\mathbf{u}}
\newcommand{\nsv}{m}
\DeclareMathOperator*{\argmax}{argmax}
\DeclareMathOperator*{\argmin}{argmin}
\renewcommand{\o}{o}
\renewcommand{\i}{i}

\usepackage{natbib}

\usepackage[noabbrev]{cleveref}

\usepackage{optidef}

\usepackage{graphicx}

\begin{document}
\makeatletter
\def\ps@pprintTitle{%
 \let\@oddhead\@empty
 \let\@evenhead\@empty
 \def\@oddfoot{}%
 \let\@evenfoot\@oddfoot}
\makeatother

\begin{frontmatter}

\title{Towards Explaining Anomalies:\\A Deep Taylor Decomposition of One-Class Models}

\author[tu]{Jacob Kauffmann}
\ead{j.kauffmann@tu-berlin.de}

\author[tu,kor,mpi]{Klaus-Robert M\"uller\corref{cor1}}
\ead{klaus-robert.mueller@tu-berlin.de}

\author[tu]{Gr\'egoire Montavon\corref{cor1}}
\ead{gregoire.montavon@tu-berlin.de}

\address[tu]{
Department of Electrical Engineering \& Computer Science, Technische Universit\"at Berlin, Marchstr. 23, Berlin 10587, Germany
}
\address[kor]{
Department of Brain \& Cognitive Engineering, Korea University, Anam-dong 5ga, Seongbuk-gu, Seoul 136-713, South Korea
}
\address[mpi]{
Max Planck Institute for Informatics, Stuhlsatzenhausweg, Saarbr\"ucken 66123, Germany
}

\cortext[cor1]{Corresponding authors}

\begin{abstract}
A common machine learning task is to discriminate between normal and anomalous data points. In practice, it is not always sufficient to reach high accuracy at this task, one also would like to understand why a given data point has been predicted in a certain way. We present a new principled approach for one-class SVMs that decomposes outlier predictions in terms of input variables. The method first recomposes the one-class model as a neural network with distance functions and min-pooling, and then performs a deep Taylor decomposition (DTD) of the model output. The proposed One-Class DTD is applicable to a number of common distance-based SVM kernels and is able to reliably explain a wide set of data anomalies. Furthermore, it outperforms baselines such as sensitivity analysis, nearest neighbor, or simple edge detection.
\end{abstract}

\begin{keyword}
one-class SVM, interpretability, deep Taylor decomposition, kernel machines, outlier detection, unsupervised machine learning.
\end{keyword}

\end{frontmatter}

\section{Introduction}
\label{section:introduction}
\begin{figure*}[ht]
\centering
\makebox[\textwidth][c]{\includegraphics[width=1\textwidth]{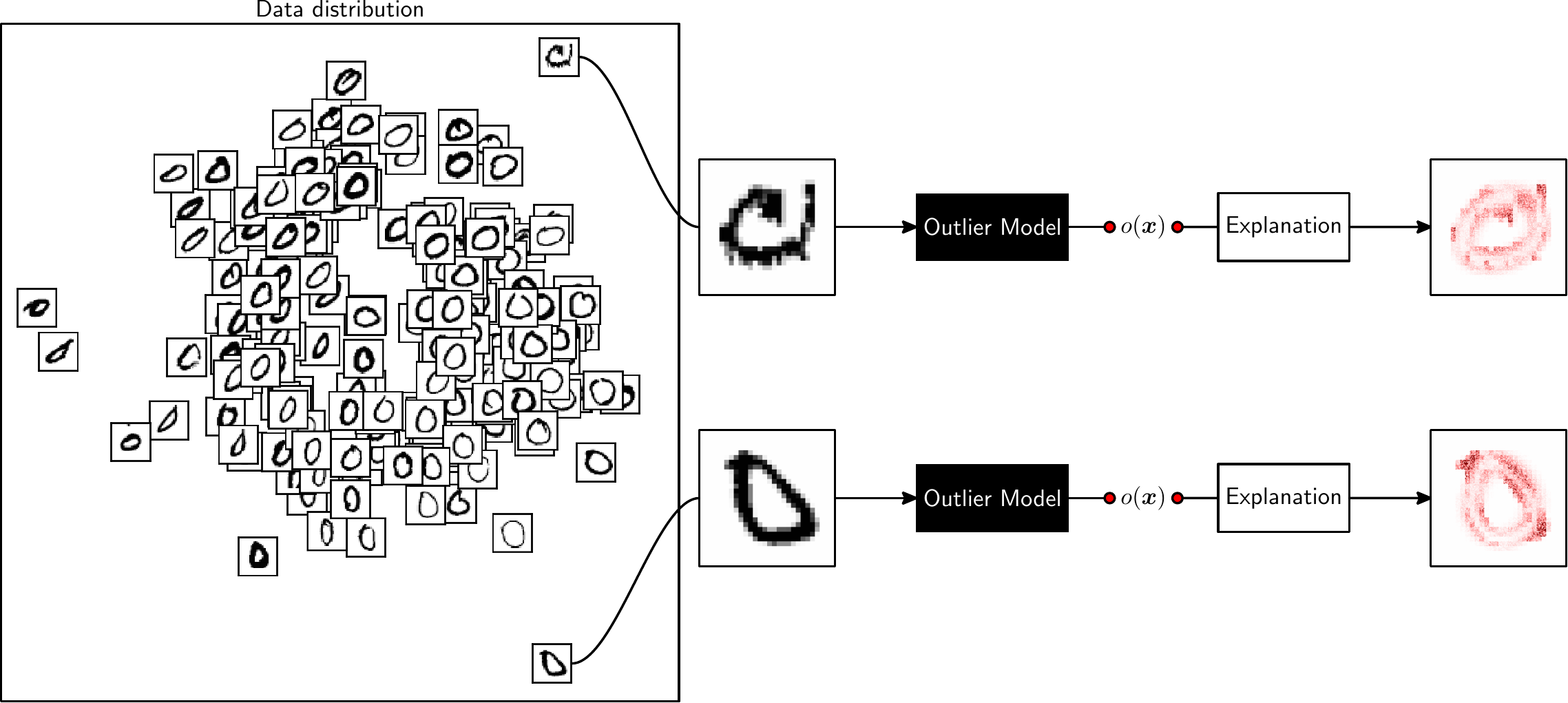}}
\caption{Illustration of the outlier detection and explanation setting. {\em Left:} Data is generated from an unknown distribution, we are for example interested in potential outliers; {\em Middle:} Unsupervised machine learning techniques estimate the data generating distribution and assign an outlier score $o(\boldsymbol{x})$ to unlikely data points; {\em Right:} Our explanation method assigns a relevance score to every input variable that reflects the contribution of input variable $x_i$ to the model decision. We apply dithering to all heatmaps for printing reliability.}
\label{figure:intro}
\end{figure*}

Novelty detection, or outlier detection, is a well-studied and well-formalized machine learning problem with numerous practical applications. One such application is intrusion detection in computer systems, where data points are typically digital messages transmitted over a network, and messages that are detected as outliers are considered likely to carry a threat \citep{DBLP:journals/tse/Denning87, DBLP:journals/jair/GoernitzKRB13}. Another application is obstacle detection in autonomous car driving \citep{hane2015obstacle}. The ability to detect outliers is also important in scientific applications, where points detected as such are intrinsically more interesting than inliers, and should therefore be given more attention \citep{Zhang2004, laurikkala-informal}. A number of techniques can be used for outlier detection
\citep{DAY1969,DBLP:journals/neco/Hinton02,Pearl:1988:PRI:534975,DBLP:conf/nips/ScholkopfWSSP99,tax2004support}.
In practice, it is not only important to be able to detect outliers and inliers with high accuracy, one would also like to be able to {\em explain} why a machine learning model considers a sample as inlier or outlier. An interpretable explanatory feedback can indeed be used by a human operator for appropriate decision making. The data point could either be considered as benign and possibly incorporated to the dataset, or appropriate action might be taken. The problem of outlier explanation is shown schematically in Figure \ref{figure:intro}. A dataset, here one class from the MNIST data set of handwritten digits, is fitted by a one-class model from which outlier scores can be obtained. These scores must then be traced back to interpretable quantities such as the input variables of the model.

Interpretability of machine learning models has received growing attention, especially in scientific applications \citep{quantum, DBLP:journals/bioinformatics/KrausBF16, sturm-jnm16, MINF:MINF201100059,Vidovic2017} and for systems that interact with humans \citep{DBLP:conf/kdd/CaruanaLGKSE15, kamarinou2016machine, 2017arXiv171108037L, DBLP:journals/corr/BojarskiYCCFJM17}. A number of generally applicable techniques for interpreting machine learning models have been proposed \cite{DBLP:journals/corr/SimonyanVZ13,erhan2009visualizing,DBLP:conf/eccv/ZeilerF14,10.1371/journal.pone.0130140,DBLP:conf/kdd/CaruanaLGKSE15}. Most of them have been developed in the context of supervised learning.

Therefore, the present work addresses the present lack of interpretability of unsupervised machines learning models, and provides a practical solution in the context of one-class SVMs for outlier detection.
We will first argue that the problems of explaining inlier and outlier decisions are qualitatively different, and need to be treated in distinct ways. Inliers will be best explained in terms of contribution of support vectors, whereas outliers will be better explained in terms of contributions of input variables. 
We propose fairly general conditions for {\em inlierness} and {\em outlierness} that can be reconciled with many common models.

Exemplarily, this will be reflected by the identification of two distinct compositions of the one-class SVM. The first one will perform a sum-pooling over similarity scores. This architecture enables the interpretation of inlierness. The second one will perform a min-pooling over distances, which provides interpretation of outlierness. In particular, we will propose in this paper a deep Taylor decomposition decomposition/integrated gradients approach \citep{DBLP:journals/pr/MontavonLBSM17,DBLP:journals/corr/SundararajanTY17}.
The proposed method can be applied to a number of outlier detection models, namely those of RBF-type \cite{DBLP:journals/vldb/KnorrNT00,tax2004support,harmeling2006outliers,bishop1994novelty,DBLP:journals/pr/Hoffmann07,DBLP:conf/sdm/YangLP09,DBLP:conf/sigmod/BreunigKNS00,tax1998outlier}. For that, the model does not have to be modified and neither re-training nor access to training data are required for the presented explanation method. Instead, only the detection model needs to be known and an appropriate measure of outlierness has to be constructed. The latter will be formally defined in \Cref{section:characterizing}. In \Cref{section:experiments}, we will show empirically that the proposed technique provides meaningful explanations.

\subsection{Related work}
A number of studies have considered the problem of outlier explanations: \citet{schwenk2014detecting} applied structured one-class SVMs to explaining anomalies in MediaCloud applications, and proposed a technique to decompose their predictions in terms of input variables for sum-decomposable kernels. We extend the previous work by proposing a Taylor-based decomposition framework applicable to various non-decomposable RBF-type kernels.
\citet{2017arXiv171110589L} use the decision of a complex outlier detection model to train a set of simple detectors that separate outliers linearly from clusters of nearby training patterns. Subsequently, the linear weights are used for interpretation of the outlier. \citet{micenkova2013explaining} heuristically remove features from detected outliers and return a subset of features that maximizes separability of the outlier from the surrounding training patterns.
These methods rely on (1) the existence of a hypothetical outlier class that is approximated by sampling in the vicinity of the supposed outliers and (2) access to the training data in the explanation stage. On the other hand, the methods are implementation invariant and model agnostic and can be applied to any outlier detection model.

We take on a different approach, where we look at the model as a mathematical function and identify marginal contributions of input variables on the produced detection score.

\section{One-Class SVM}\label{section:oneclass}

In one-class learning, we are trying to separate patterns that are generated by one common distribution from the rest of the input domain. 
\citet{DBLP:conf/nips/ScholkopfWSSP99} proposed the one-class SVM as an algorithm that learns the tails of a high-dimensional distribution, which is sufficient for the separation task. For a set of training data $\x_1,\dots,\x_n\in\mathcal{X}$ and some feature map $\Phi\colon\mathcal{X}\rightarrow\mathcal{F}$, the primal one-class SVM problem takes the form
\begin{mini*}
{\substack{\boldsymbol{w}\in\mathcal{F},\rho\in\mathbb{R},\boldsymbol{\xi}\in\mathbb{R}^n}}
{\frac{1}{2}\|\boldsymbol{w}\|^2 - \rho + \frac{1}{\nu n}\sum_{i=1}^{n}\xi_i}
{\label{optimize_ocsvm}}
{}
\addConstraint{\langle\boldsymbol{w},\Phi(\boldsymbol{x}_i)\rangle~}
{\geq \rho - \xi_i,\quad\forall i = 1,\ldots,n}{}
\addConstraint{\xi_i ~}
{\geq 0\label{constraint_convex_hull}}{}
\end{mini*}
where $\nu\in[0, 1]$ controls the fraction of outliers that are extracted by the model \cite{DBLP:conf/nips/ScholkopfWSSP99}.
Given an explicit map $\Phi$ with interpretable features (e.g.\ BoW or pixel intensities), the one-class SVM is readily interpretable in feature space $\mathcal{F}$ by means of the linear weight vector $\boldsymbol{w}\in\mathcal{F}$.
For RBF-kernels, the optimization is performed in the dual formulation, which does not provide an explicit representation of the weight vector, but a set of Lagrangian multipliers, taken as coefficients of radial basis functions, centered at support vectors.
Let $k\colon\mathbb{R}\rightarrow\mathbb{R}$ be a RBF-kernel that acts on the distance of two points and produces large output for patterns that are similar and small output for distinct patterns. 
The one-class SVM extracts a small set of support vectors $\sv_1,\ldots,\sv_\nsv$ with $\nsv \leq n$ from the training set together with coefficients $\alpha_1,\ldots,\alpha_\nsv$ such that
\begin{align*}
g(\x) = \sum_{j=1}^\nsv \alpha_j\,k(\|\x - \sv_j\|)
\end{align*}
is large for data points $\x\in\mathcal{X}$ that are {\em typical} in terms of the training data and the chosen similarity measure $k$. For anomalous points, $g(\x)$ will be small.

\section{Inlierness and Outlierness}
\label{section:characterizing}

{
Having introduced the one-class SVM, we now take a more abstract look at the problem. We will characterize inliers and outliers by answering: (1) what is an appropriate compositional structure for these two quantities, and (2) how should inlierness and outlierness be quantified. Our answers to these questions will provide a theoretical basis for the design of our explanation method described in Sections \ref{section:kerneltaylor} and \ref{section:deeptaylor}.
}

\subsection{Modeling Inlierness and Outlierness}

\begin{figure*}[ht]
\centering
\makebox[\textwidth][c]{\includegraphics[width=1\textwidth]{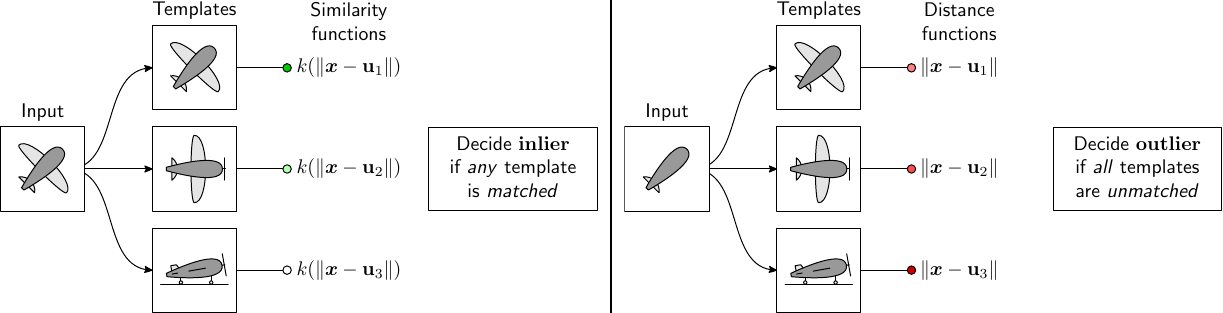}}
\caption{The compositional structure of inlier and outlier decision is substantially different. {\em Left:} Inlier decisions are characterized by a max-pooling over similarity activations, {\em right:} outlier decisions are a min-pooling over distance activations.}
\label{fig:inoutcomic}
\end{figure*}

Any complex prediction task requires a set of function classes to choose from. These functions are preselected based on some prior knowledge about the problem, and incorporate properties such as linearity, smoothness, and more general types of equivariances or invariances. Practically, these function classes can be implemented by a model which can be, for example, a composition of multiple layers.

The compositional structure of the model differs substantially between the types of prediction tasks. For example, a model that detects ``airplanes'' in an image will typically consist of multiple detectors that test for the presence of an airplane at various locations in the image. The detection decision can be expressed as: {\em ``Decide `airplane' if \underline{any} airplane template is matched.''} An appropriate architecture for this problem would therefore be a collection of similarity functions in the first layer, followed by a max-pooling operation in the second layer. This structure of the prediction function is prototypical for state-of-the-art classification architectures such as the deep convolutional neural network, where detection layers are interleaved with max-pooling layers \citep{boureau2010theoretical}.

Max-pooling architectures are also particularly suitable for the problem of detecting inliers. A first layer will detect the similarity to every individual airplane in the data, and a second layer will retain the maximum similarity scores obtained in the previous layer. Here, each airplane detector measures the similarity to an airplane or a group of airplanes in the data. The inlier decision can in that case also be expressed as {\em ``Decide `inlier' if \underline{any} airplane template is matched.''}, i.e.\ in the same way as for the detection task. An appropriate composition of the inlierness function is therefore of type $\i(\x) = \max_j k(\|\x-\sv_j\|)$, where the first layer maps the input data to the similarity function scores, and the second layer applies some max-pooling operation or a soft variant of it.\footnote{Typical soft variants of max-pooling are the sum, $\ell_p$-norm, arithmetic mean, or log-sum-exp. Henceforth, we refer to these functions as soft max-pooling.} This structure is visualized in \Cref{fig:inoutcomic} (left).

On the other hand, if we were using the same max-pooling approach for outlier detection, one would need to build as many detectors as there are possible inputs without an airplane. There is an exponential number of them. Instead, the outlier detection problem is better expressed as follows: {\em ``Decide `outlier' if \underline{all} airplane templates are unmatched.''} In that case, the first layer models the level of distance functions, and the second layer becomes a min-pooling operation. An appropriate composition of the outliereness function will be of type $\o(\x) = \min_j \|\x-\sv_j\|$, where the first layer maps the input data to the distances, and the second layer applies some min-pooling operation or a soft variant of it. This new structure is visualized in \Cref{fig:inoutcomic} (right).

\subsection{Quantifying Inlierness and Outlierness}
\label{sec:quantifying}
In problems such as classification and regression, the output of the model can be readily interpreted, e.g.\ as the probability of membership to a given class, or as the expected value of the target variable respectively. When using, e.g.\ a one-class SVM, such interpretation is not obvious: The discriminant function $g(\x)$ does provide an ordering from the most to the least anomalous point (cf.\ \citet{harmeling2006outliers}), however, it only answers which of two data points is most anomalous, and not the absolute level of anomaly of a given data point. We propose the following axiomatic definitions for inlierness and outlierness, and then briefly discuss how common machine learning outlier detection models fulfill or violate these definitions:
\begin{definition}
A {\em measure of inlierness} $i\colon\mathcal{X}\rightarrow\mathbb{R}$ must fulfill the following two conditions for all $\x\in\mathcal{X}$: 
\begin{enumerate}
\item It is bounded by zero and some positive number $u$: $0\leq i(\x) \leq u$ and
\item It converges asymptotically to zero: $\lim_{t\rightarrow\infty}i(t\x) = 0$.
\end{enumerate}
\label{def:inlier}
\end{definition}
For example, the Gaussian mixture model, which is sometimes used for inlier/outlier detection (e.g.\ \cite{tax1998outlier}), associates to each input point a probability score representing the likelihood of that point being generated from the underlying distribution \citep{Bishop:2006fk}. This probability score is bounded between $0$ and $1$ and converges to $0$ when moving away from the data. Thus, these probability scores fullfill our definition of inlierness. Similarly, the discriminant function of the one-class SVM with RBF kernel is upper boun\-ded by the kernel bound, and converges to zero as we move away from the data.

These quantities are however not suitable as an outlierness model:
they asymptote to $0$ as $\x$ moves away from the data, which does not captures the fact that the degree of outlierness continues to increase. Outlierness is instead better defined by the following set of axioms:
\begin{definition}
A function $\o\colon\mathcal{X}\rightarrow\mathbb{R}$ is called {\em measure of outlierness} if it fulfills the following conditions for all $\x\in\mathcal{X}$:
\begin{enumerate}
\item It is lower bounded by zero: $0\leq o(\x)$ and
\item\label{def:outlier-norm} It converges asymptotically with the Euclidean norm
$
\frac{\o(t\x)}{\|t \x\|^q} = c,$ for some $q \geq 1$ and some $c>0$.
\end{enumerate}
\label{def:outlier}
\end{definition}
To reflect the Euclidean geometry of the input space, the norm in the denominator will be assumed to be a $\ell_2$-norm. Example of functions that satisfy \Cref{def:outlier} are the distance to the mean, or the neg-log-likelihood under an isotropic probability distribution, e.g.\ $\mathcal{N}(\boldsymbol{\mu},\sigma^2I)$. These function are typically used in machine learning for measuring error.

As a counter example, the neg-log-likelihood of a general Gaussian distribution $\mathcal{N}(\boldsymbol{\mu},\Sigma)$ learned from the data does not satisfy \Cref{def:outlier}: The latter is indeed not suitable for measuring outlierness, as the learned covariance $\Sigma$ overrides the natural metric of input space on which the outlier decision should be based.

Having defined inlierness and outlierness, we now provide measures for the one-class SVM of interest in this paper. These measures are based on the discriminant $g(\x)$ defined above. In general, there may be more than one measure of inlierness or outlierness, and we shall here apply a principle of parsimony.
\paragraph{Exponential kernels}
The first class of kernels we consider are exponential kernels which can be parametrized as
\[
k(\|\x-\x'\|) = \exp\left(-\frac{\|\x-\x'\|^q}{q \cdot \sigma^q}\right).
\]
Parameter $\sigma$ is the bandwidth of the kernel. For $q=1$, the kernel is called {\em Laplacian}, for $q=2$ {\em Gaussian kernel}.
The simplest measures of inlierness and outlierness that satisfy Definitions \ref{def:inlier} and \ref{def:outlier} would be:
\begin{align}
i(\x) = g(\x), && o(\x) = -\log(g(\x)).
\label{eq:measures-exponential}
\end{align}
A proof that the outlierness meets \Cref{def:outlier} can be found in \ref{sec:out_conform}.
\paragraph{$t$-Student kernels}
The second class of kernels that we consider are $t$-Student kernels:
\[
k(\|\x-\x'\|) = \frac{1}{a + \|\x-\x'\|^q}.
\]
The parameter $a$ is positive and
often set to 1. When the norm is also scaled by a bandwidth, the kernel is also referred to as Cauchy kernel.
Inlierness and outlierness will be measured by the following functions:
\begin{align}
i(\x) = g(\x), && o(\x) = \frac{\nsv}{g(\x)}.
\label{eq:measures-student}	
\end{align}
A proof for the agreement of $o(\x)$ with \Cref{def:outlier} is in \ref{sec:out_conform}.

\section{Explaining Machine Learning Decisions}
\label{section:background}

In this section, we review several techniques to explain the predictions of a machine learning classifier in terms of input variables. Let $\x \in \mathbb{R}^d$ be an input example and $f(\x) \in \mathbb{R}$ be its prediction, where $f$ is a function learned from the data. The goal of an explanation is to assign a relevance score $R_i$ to each feature $x_i$, that reflect the importance of that feature for the prediction.

\paragraph{Sensitivity Analysis} The simplest technique for explanation is to attribute relevance to the input variables to which the prediction is locally most sensitive \citep{DBLP:conf/iscas/ZuradaMC94, Gevrey2003, DBLP:journals/jmlr/BaehrensSHKHM10}. That is, for a given prediction, we define the importance score for each input variable $i$ as:
$$
R_i = \left(\frac{\partial f}{\partial x_i}\Big|_{\x}\right)^2,
$$
that is, the squared locally evaluated partial derivatives.
A limitation of sensitivity analysis is that it is an explanation of the function variation rather than of the function value.
Considering a single distance norm $\|\x - \sv_j\|$, we observe that the gradient does not grow with the distance, implying that sensitivity analysis does not capture the amount of outlierness that a pattern holds. Another observation is that the gradient vanishes between modes of the data, imposing zero importance to variables that occupy a local maximum of outlierness, when measured with sensitivity analysis.
The aforementioned weaknesses in sensitivity have led to the development of more precise explanation techniques, which we will take up in the following.

\paragraph{Simple Taylor Decomposition}
Taylor decomposition \cite{Bazen2013,10.1371/journal.pone.0130140} seeks to determine the importance of input variables for a certain prediction $f(\x)$ by performing an expansion of the function $f$ at a certain reference point $\widetilde \x$:
\begin{align*}
f(\x) = 
{\underbrace{\vphantom{\sum_{i=1}^d} f(\xroot)}_{(1)}} &+ 
{\underbrace{\sum_{i=1}^d \frac{\partial f}{\partial x_i}\Big|_{\xroot}\cdot(x_i - \widetilde{x}_i)}_{(2)}} \nonumber + 
{\underbrace{\vphantom{\sum_{i=1}^d}\varepsilon.}_{(3)}}
\end{align*}
It then identifies as importance for a given variable the various terms of the expansion that are bound to it. In the equation above, (1) is the function value at the reference point, (2) contains linear contributions, (3) contains all higher-order terms, including interdependence relations between input variables. Simple Taylor decomposition focuses on the term (2), where the summands are bound to a given input variable. Thus, we define the relevance scores $(R_i)_i$ for the prediction $f(\x)$ as:
$$
R_i = \frac{\partial f}{\partial x_i}\Big|_{\xroot}\cdot(x_i - \widetilde{x}_i).
$$
In our analysis, we will also choose functions and reference points such that the term (1) is zero, i.e.\ contains no information on the models prediction and (3) is small. In that case, we obtain the relevance conservation property $\textstyle \sum_{i=1}^d R_i \approx f(\x)$, which guarantees that the explanation matches in magnitude the amount of predicted inlierness or outlierness. A limitation of simple Taylor decomposition is the need to find a root point $\xroot$ in the vicinity of $\x$, which can be time-consuming.
Further, the reference point might {\em jump} to a different mode as the input pattern moves from one mode to another mode of the distribution. This may cause two nearly equivalent data points with nearly equivalent predictions to receive a different explanation.
Stated differently, the explanation as a function of $\x$ is discontinuous.

\paragraph{Integrated Gradients}
Another approach for setting importance scores of inputs to a prediction has been proposed by \citet{DBLP:journals/corr/SundararajanTY17}. For some reference point $\widetilde\x$, a prediction is explained by summing over a finite number of small steps of first order simple Taylor decompositions between the input $\x$ and the reference point $\widetilde\x$.
In the limit, the attribution can be written in terms of the integral
\begin{align*}
f(\x) = f(\widetilde\x) + \sum_{i=1}^d\underbrace{(x_i - \widetilde{x}_i)\cdot\int_0^1\frac{\partial f}{\partial x_i}\Big|_{\widetilde\x+\alpha(\x-\widetilde\x)}d\alpha}_{R_i}
\end{align*}
which typically has to be evaluated numerically, but may also have an analytical solution for simpler models. Like for simple Taylor decomposition, one needs to choose an appropriate reference point. The advantage of integrated gradients is the absence of second- and higher-order residual terms.
In \Cref{section:deeptaylor-input} we apply the method to convex functions of which the integral has an analytical solution.

\paragraph{Deep Taylor Decomposition}
Deep Taylor decomposition (DTD) is a method for decomposing the prediction of a neural network on its input variables \cite{DBLP:journals/pr/MontavonLBSM17}. The decomposition is obtained by propagating the model output into the neural network graph by means of redistribution rules, until the input variables are reached. As such, it belongs to a broader class of propagation techniques \cite{10.1371/journal.pone.0130140,DBLP:conf/cidm/LandeckerTBMKB13,DBLP:journals/corr/ShrikumarGK17,DBLP:journals/corr/Zhang0BSS16}.
A distinctive feature of DTD is that the propagation rules are derived from a Taylor decomposition performed at each neuron of the network.

The decomposition process starts from the top neuron, whose activation is redistributed into relevance scores of neurons in the previous layer. The previous layer relevance scores are then expressed as a function of the activations of the layer before, which enables another step of redistribution. The Taylor decomposition process is iterated from the top layer down to the input layer where the decomposition in each layer has a closed form for known compositions. The procedure leads ultimately to a relevance score for each input variable. Like for the forward pass or standard gradient propagation, DTD can be quickly computed in $\mathcal O(\# \text{connections})$.

The original DTD method uses Taylor decomposition as a unit of explanation at each neuron. However, our adaptation of DTD in the context of one-class SVM leads to the observation that, for certain neuron types, e.g.\ mapping on the kernel basis function, this unit of explanation can be advantageously substituted by other analyses such as integrated gradients. Overall, the method we present in this paper generalizes DTD to a ``deep decomposition'' where we use standard Taylor decomposition or integrated gradients as unit of explanation at various layers.

\paragraph{Other methods} A number of other methods have been proposed for explanation: It includes methods based on locally sampling the decision function \cite{DBLP:conf/kdd/Ribeiro0G16}, local perturbations \cite{DBLP:conf/eccv/ZeilerF14,DBLP:journals/corr/ZintgrafCAW17}, other types of propagation techniques \cite{DBLP:conf/eccv/ZeilerF14,DBLP:journals/corr/SpringenbergDBR14}, as well as explanation methods supported by specific choices of achitectures \cite{DBLP:conf/cvpr/ZhouKLOT16,DBLP:conf/kdd/CaruanaLGKSE15}.
\section{Explaining Inlierness}
\label{section:kerneltaylor}

\begin{figure*}
\centering
\includegraphics{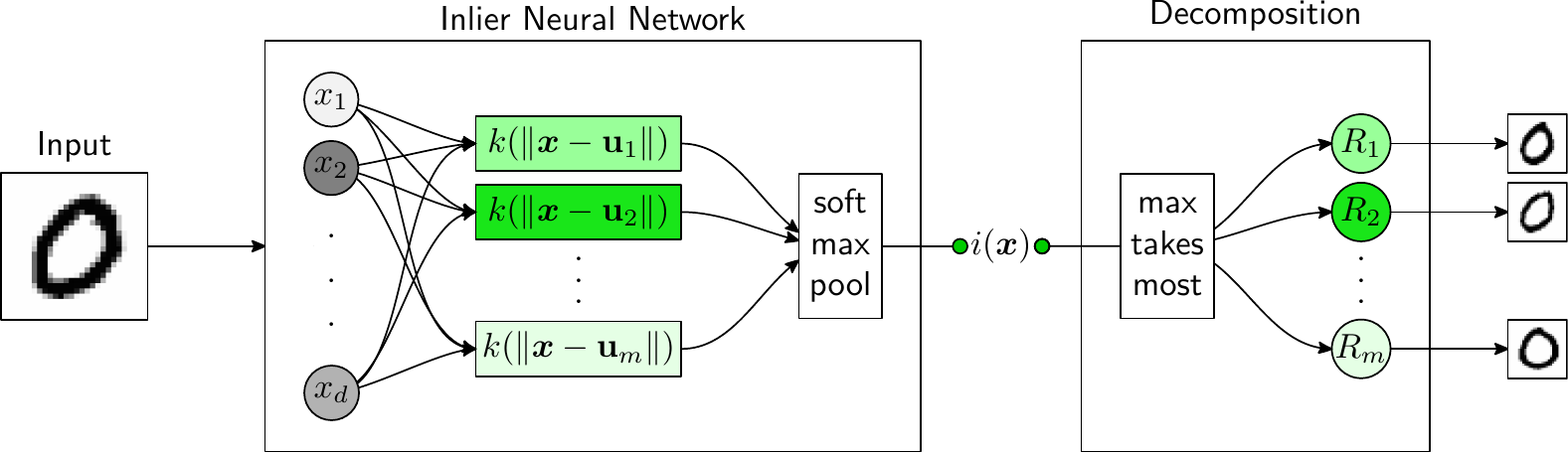}
\caption{{ Neural network equivalent of the one-class SVM for inlier detection, and the relevance redistribution from the top layer to the intermediate layer.}}
\label{figure:flow-inlier}
\end{figure*}
In this section, we present the decomposition of the measure of inlierness $\i(\x)$ defined in \Cref{sec:quantifying}. As it was argued in Section \ref{section:characterizing}, the inlierness is best modeled by a detection-max-pooling architecture. 
Such architecture is common in convolutional neural networks, where max-pools are composed of outputs from different detectors that were applied to the same lower level features.
A two-layer neural network that implements the measures of inlierness is given by:
\begin{align*}
&\text{layer 1:} & s_j &= \alpha_jk(\|\x-\sv_j\|) & \text{(detection)},\\
&\text{layer 2:} & \i &= \sum_{j=1}^\nsv s_j& \text{(pooling)},
\end{align*}
where the first layer are the weighted similarities to the support vectors measured by the kernel, and the second layer performs a sum-pooling, which can be viewed as a soft variant of max-pooling. Deep Taylor decomposition applies as a first step the decomposition of the output $\i(\x)$ on the first-layer activations $(s_j)_j$ that we call ``effective similarities'' due to the weighting term $\alpha_j$. The two-layer architecture and the process of relevance redistribution from the top layer to the intermediate layer is shown in \Cref{figure:flow-inlier}.

The input (e.g.\ a handwritten digit) is first propagated into the neural network, to compute the inlier score. Then, this score is redistributed from the top layer to the hidden layer, which gives a decomposition of inlierness in terms of support vectors. Technically, we perform a Taylor expansion of the inlier as a function of the hidden layer activations $\i((s_j)_j)$. Relevance scores are then given by:
\begin{align}
R_j &=  \sum_j \frac{\partial \i}{\partial s_j}\Big|_{(\widetilde s_j)_j} (s_j - \widetilde{s}_j)
\label{eq:taylor-sumpooling}
\end{align}
Due to the linearity of the sum-pooling function, there is no second order term. In order to satisfy the conservation property $i = \sum_j R_j$, we further need to have $\i((\widetilde s_j)_j) = 0$, i.e.\ we need to perform the Taylor expansion at a root point of the function. Here, we choose the root point $(\widetilde{s}_j)_j = (0)_j$, because it is the only admissible root point in the space of activations. The deep Taylor decomposition method \cite{DBLP:journals/pr/MontavonLBSM17} does not require the root point to have a pre-image in the lower-layer, however, in this particular case, one can still interpret the segment $[(s_j)_j,(0)_j]$ as moving in some direction orthogonal to the data manifold in the input domain.

Injecting the root point $(0_j)_j$ in \cref{eq:taylor-sumpooling} gives the relevance score:
\begin{align*}
R_j = s_j
\end{align*}
That is, the relevance of support vector $\sv_j$ corresponds to its hidden neuron activation $s_j$. This operation can be interpreted as a ``max-takes-most'' redistribution.

We now ask if it is sufficient for explanation to stop at this layer, or if relevance should be further propagated to the input variables. For this, consider the simplest inlier model $i(\x) = \alpha \cdot k(\|\x-\sv\|)$ composed of a single support vector $\sv$. Consider the most inlier point $\x^\star = \argmax_{\x} i(\x) = \sv$. At this location, it is easy to conclude that $\sv$ has contributed to the inlierness of $\x^\star$, however, because the kernel is RBF and $\x^\star$ lies at the maximum, it is impossible to assign a directional explanation for such inlinerness. Indeed, $i(\x)$ looks exactly the same along each direction. Based on this prototypical example, one concludes that explanations for inlierness are better given in terms of support vectors than input directions.

\section{Explaining Outlierness}
\label{section:deeptaylor}
As it was discussed in Section \ref{section:characterizing}, outlier detection is more naturally described as a min-pooling over local distances.
Unlike explanation of inliers, the analysis here will depend on the choice of kernel. For each family of kernel, one needs to find a suitable model composition, and appropriate root points for the explanation. In this section, two classes of kernels are considered. These kernels are frequently encountered in practical applications.

\subsection{$t$-Student Kernels}
\label{section:deeptaylor-student}

The first kernel we focus on is the generalized $t$-Student kernel given by $k(\|\x-\x'\|) = (a + \|\x-\x'\|^q)^{-1}$. We compute the one-class SVM discriminant $g(\x) = \sum_j \alpha_j k(\|\x-\sv_j\|)$, and apply the measure of outlierness $\o(\x) = \nsv g(\x)^{-1}$ proposed in \Cref{sec:quantifying} for this kernel. The measure of outlierness $\o(\x)$ can be implemented by the following two-layer neural network (see \ref{sec:nnrepproof} for a proof):
\begin{align*}
&\text{layer 1:} & h_j &= \frac{1}{\alpha_j} \cdot (a + \|\x - \sv_j\|^q) & \text{(detection)},\\
&\text{layer 2:} & \o &= \mathrm{H}((h_j)_j) & \text{(pooling)},
\end{align*}
The first layer can be interpreted as a mapping to the effective distances $h_j$ from each support vector. By effective distance, we mean the distance as perceived by the data point $\x$, i.e.\ modulated by the support vector coefficients $\alpha_j$. The second layer computes the harmonic mean $\mathrm{H}$ which implements a soft min-pooling.

We would now like to redistribute the output $\o$ to the lower-layer. We let $\o$ depend on the hidden layer activations so that a Taylor decomposition can be performed on the previous layer. Specifically, we choose the root point $(\widetilde{h}_j)_j = (0)_j$ and perform a first order Taylor decomposition at that point. It can be shown that higher order terms sum to one in the Taylor expansion for that root (see \ref{sec:linproof}). Relevance scores are given by:
\begin{align*}
R_j &= \frac{\partial \o}{\partial h_j} \Big|_{(h_j)_j = (\widetilde h_j)_j} \cdot (h_j - \widetilde h_j)\\
&= \frac{m \cdot \left(\frac{1}{h_j}\right)^2}{\left(\sum_{j'=1}^{m} \frac{1}{h_{j'}}\right)^2} \cdot h_j \stepcounter{equation}\tag{\theequation}\label{eq:svr_student}\\
&= h_j \cdot c_j \qquad \text{with} \quad c_j = \frac1m\cdot\mathrm{H}\bigg(\bigg(\frac{h_{j'}}{h_j}\bigg)_{j'}\bigg)^2.
\end{align*}
where $h_j$ is the first-layer activation representing effective distances, and $c_j$ is a factor that only retains support vectors that are active in the min-pooling operation (i.e.\ those with the lowest effective distance). In the input domain, $c_j$ can be interpreted as a localization term. A large relevance score $R_j$ is therefore the result from a large effective distance $h_j$, but low in comparison to other effective distances $(h_{j'})_{j'}$ in the pool.
In \ref{sec:linproof}, we show that the decomposition is conservative, i.e. $\sum_j R_j = o$.
In \Cref{section:deeptaylor-input} we will show how to redistribute $R_j$ to the input layer.

\subsection{Exponential Kernels}
\label{section:deeptaylor-exponential}

In this section we consider the family of kernels of type
$k(\|\x-\x'\|) = \exp\left(-\left(\|\x-\x'\|\slash\sigma\right)^q\slash q \right)$. Unlike the kernels of Section \ref{section:deeptaylor-student}, this family of kernel implements stronger locality.
The Laplacian and Gaussian kernels are special cases for $q=1$ and $2$ respectively. Like in the previous section, we compute the SVM discriminant $g(\x) = \sum_j \alpha_j k(\|\x-\sv_j\|)$, however, we apply a different measure of outlierness, $\o(\x) = - \log g(\x)$, proposed in \Cref{sec:quantifying} for this kernel. The function $o(\x)$ can be mapped to the following two-layer neural network (proven in \ref{sec:nnrepproof}):
\begin{align*}
&\text{layer 1:} & h_j &= -\log(\alpha_j) + \frac1q \left(\frac{\|\x - \sv_j\|}{\sigma}\right)^q & \text{(detection)}\\
&\text{layer 2:} & \o &= -\mathrm{LSE}(-(h_j)_j) & \text{(pooling)}
\end{align*}
a set of radial basis distance functions followed by a flipped log-sum-exp computation which implements a soft of min-pooling. 
We let the neural output depend on the hidden layer, and choose the root point $(\widetilde{h}_j)_j = (h_j)_j - (o)_j$, i.e.\ we substract the output of the model to each dimension of the vector of activations. Relevance scores on the hidden layer are obtained by Taylor decomposition:
\begin{align*}
R_j &= \frac{\partial \o}{\partial h_j} \Big|_{(h_j)_j = (\widetilde h_j)_j}^\top \cdot (h_j - \widetilde h_j)\\
&= \frac{\exp(-h_j)}{\sum_{j'=1}^m \exp(-h_{j'})} \cdot (-\mathrm{LSE}((-h_{j'})_{j'}) \stepcounter{equation}\tag{\theequation}\label{eq:svr_exponential}\\
&= (h_j + \varepsilon_j) \cdot p_j \quad \text{with} \quad p_j = \frac{\exp(-h_j)}{\sum_{j'} \exp(-h_{j'})} \\[1mm]
& \qquad \qquad \qquad \qquad \text{and} \quad ~\varepsilon_j = - \mathrm{LSE}(-(h_{j'}-h_j)_{j'}).
\end{align*}
One can also show that this decomposition is conservative (see \ref{sec:linproof}).
\subsection{Redistribution on the Input Layer}
\label{section:deeptaylor-input}

\begin{figure*}
\centering
\makebox[\textwidth][c]{\includegraphics[width=1\linewidth]{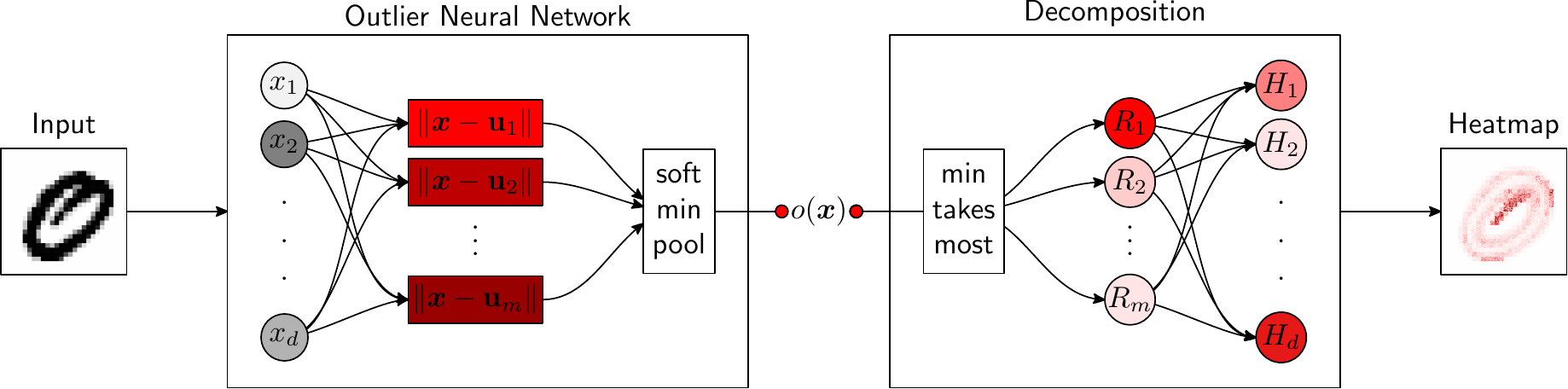}}
\caption{Neural network equivalent of the one-class SVM for outlier detection, along with the relevance propagation procedure to determine pixel-wise contributions to outlierness.}
\label{figure:flow-outlier}
\end{figure*}

In the inlier detection case, it was sufficient to perform redistribution on the domain of support vectors. We argue that explaining an outlier in terms of support vectors does not provide much interpretability. Take a prototypical outlier, which is very far from the data. From this distance, two distinct support vectors will look very similar, and the main information about outlierness is not contained in which point it is the closest from, but in the distance and direction between the outlier and the data. Thus, motivated by this prototypical-case argument, we now look at how to backpropagate the outlier explanation $(R_j)_j$ one layer below onto the input domain.

In Sections \ref{section:deeptaylor-student} and \ref{section:deeptaylor-exponential}, support vector relevance was given by $R_j = h_j \cdot c_j$ and $R_j = (h_j + \varepsilon_j) \cdot p_j$. Redistribution on the input domain requires to express $R_j$ as a function of $\x$. We will first show that $c_j$, $\varepsilon_j$ and $p_j$ are approximately constant: When $\forall_{j' \neq j} h_j\slash h_{j'} \approx 0$, i.e.\ when support vector $\sv_j$ dominates locally, then $c_j,p_j \approx 1$, and $\varepsilon \approx 0$. Furthermore, $c_j$ is constant under any rescaling of activations $(h_j)_j$, and $p_j, \varepsilon_j$ are constant under any increment of activations by a constant value. A proof for the invariances can be found in \ref{sec:constant}. These transformations also describe the path where we look for the root point. Thus, considering these terms as effectively constant, $R_j$ can be modeled locally as an affine transformation of activations, which are themselves an affine transformation of distances. We write:
$$
R_j = C_j\|\x - \sv_j\|^q + D_j
$$
where $C_j > 0$ and $D_j \in \mathbb{R}$ are constant. This quantity is redistributed on the input dimensions by means of integrated gradients \cite{DBLP:journals/corr/SundararajanTY17}.
A detailed derivation of integrated gradients of $R_j$ can be found in \ref{sec:dec_rj}.

The attribution on the input variables is given in vector form by
\begin{align*}
\mathbf{R} &= \sum_{j=1}^m(\x - \xroot_j)\odot\int_0^1\frac{\partial R_j}{\partial \x}\Big|_{\xroot_j+t(\x-\xroot_j)}dt \\
&= \sum_{j=1}^m\left[\frac{\x-\sv_j}{\|\x-\sv_j\|}\right]^2\cdot(R_j-D_j^+).\stepcounter{equation}\tag{\theequation}\label{equ:inputrel}
\end{align*}
where like in the original paper \cite{DBLP:journals/pr/MontavonLBSM17} we have summed over relevance received from all higher-layer units. The expression $D_j^+$ denotes $\max(0, D_j)$ and the integral is the vector of individual integrals of $\partial R_j\slash\partial x_i$.

The whole process of layer-wise redistribution from the top layer down to the input layer is shown in Figure \ref{figure:flow-outlier}. The data point (e.g.\ a handwritten digit) is given as input to the neural network. The network implements the outlier function as a soft min-pooling over support vector distances. The outlier score obtained at the output of the network is redistributed using deep Taylor decomposition: it is first redistributed using Taylor decomposition on the support vectors, and then further propagated to the input domain using integrated gradients.

\section{Extension for Sequential Data}\label{sec:seq}

When applied to sequential data such as images or time series, one-class models based on RBF kernels become affected by the curse of dimensionality. Thus, it is sometimes preferable to apply these models to small sequences or patches of the input \cite{DBLP:conf/cvpr/LiPT05,li2005learning,DBLP:conf/iccv/FromeSSM07}. The scores computed for all patches are then pooled to compute a global score for the sequence. Let $(i_t)_t$ and $(o_t)_t$ be the inlier and outlier scores associated to a collection of patches or segments taken from the input sequence. One measure of outlierness that satisfies Definition \ref{def:outlier} is obtained by summing all outlier scores, thus forming a third layer of representation:
\begin{align*}
&\text{layer 3:} & O &= \textstyle \sum_{t=1}^T o_t & \text{(pooling)}.
\end{align*}
This composition resembles the max-pooling layer 2 from \Cref{section:kerneltaylor}. Choosing the root $(\widetilde{o}_t)_t = (0)_t$ consists therefore of a max-take-most redistribution on the spatial locations, {$R_t = o_t$},
from where on we proceed as explained in \Cref{section:deeptaylor}, {by first redistributing location relevance to support vectors, $R^t_j$ by \Cref{eq:svr_student} or \Cref{eq:svr_exponential} and perform a final redistribution on input variables by \Cref{equ:inputrel}.
}
\section{Experiments}
\label{section:experiments}

\begin{figure*}
\centering
\includegraphics[width=1\textwidth]{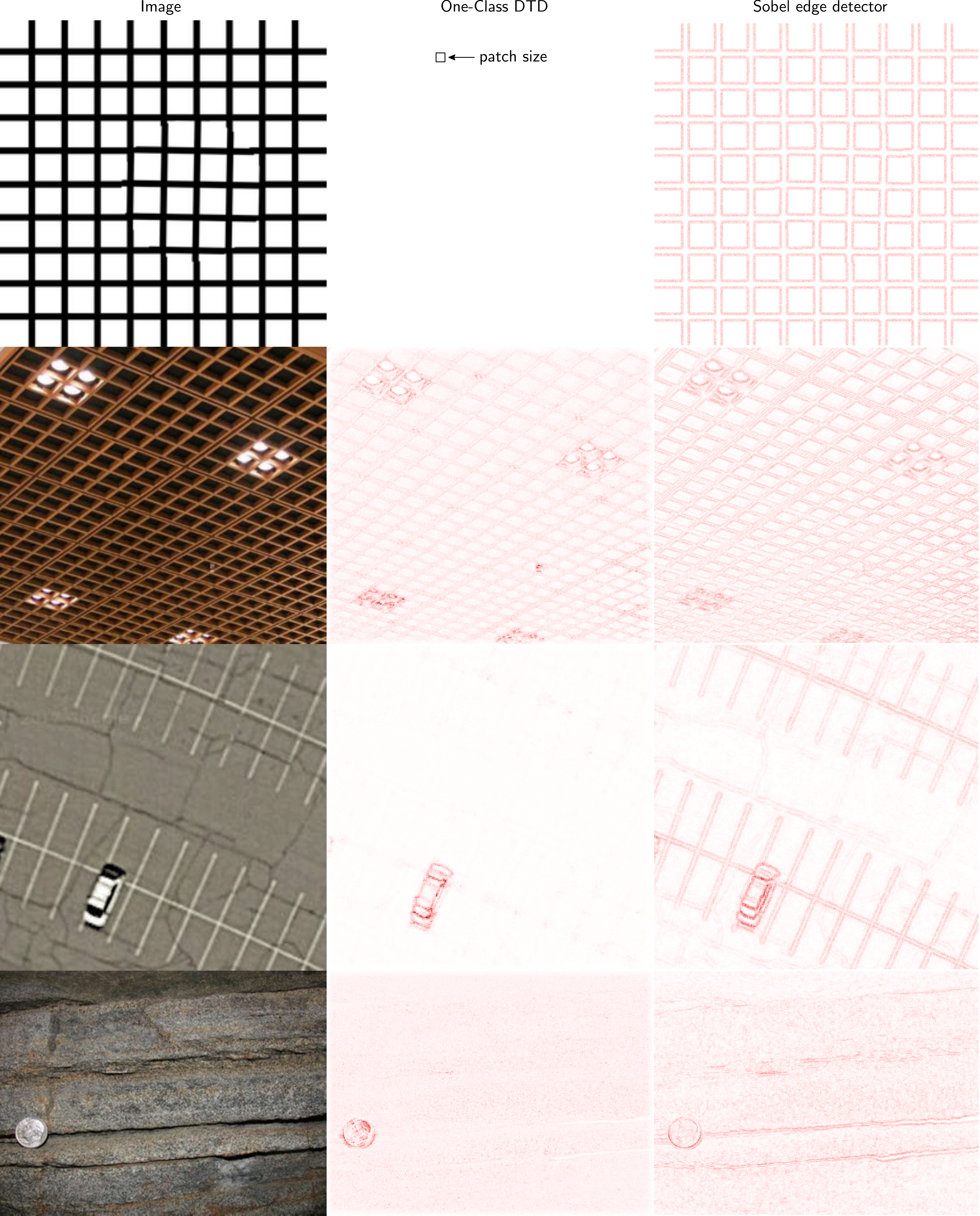}
\caption{A One-Class SVM is trained on small $7\times 7$ patches of the very image itself. Parameter $\nu=0.1$ is set to allow at most 10\% outliers. Images from a texture data set \citep{cimpoi14describing} (row one, two and four) and PatternNet \citep{DBLP:journals/corr/ZhouNLS17}; top image is altered by us. For every image, we show {\em Left:} input image; {\em Middle} decomposition of one-class SVM; {\em Right} Sobel filter for reference. All images were resized to 256 pixels width.}
\label{fig:heatmap_example}
\end{figure*}

We first test our deep Taylor decomposition (DTD)-based method for outlier explanation on large images, where we use the sequential model of Section \ref{sec:seq}. \Cref{fig:heatmap_example} shows heatmaps for images taken from various image datasets. These heatmaps are compared to a simple baseline edge detector.

All models are trained on $7\times 7$ patches from the single image itself, thus heatmaps should highlight unusual statistics in the image. The function that the model implements depends solely on model parameters (1) fraction of outliers $\nu$, here chosen as 0.1, (2) degree of Euclidean distance $q$, here set to 2, (3) kernel bandwidth $\sigma$ chosen as 0.1 quantile of one-nearest-neighbor distances for the exponential kernel\footnote{a variation of the heuristic from \citet{easy_kernel_width}} and (4) the patch size. Having these parameters fixed, the one-class SVM has a unique solution and explanation. Examples in \Cref{fig:heatmap_example} are generated with a Gaussian kernel. We rescale images to a common width of 256 pixels and apply anti-aliasing in the rescaling, because we observed that the method is sensitive to aliasing artifacts.

The One-Class DTD grounds anomalies to individual pixels. 
The first row in \Cref{fig:heatmap_example} shows a modified image from the class ``grid'' from the Describable Textures Dataset \citep{cimpoi14describing}. We perturbed the clean grid by a circle in the grid that is invisible to the human eye. The outlier edges that are due to our modification are indeed discovered by our method, as we can see in the heatmap right next to the grid image. The Sobel edge detector is not able to detect these special edges.
The second image has a small defect in the middle of the lower-right quadrant. It is not obvious that this defect can be detected in the presence of other distractions, like the lamps, that are detected as well.
The first three images show that our One-Class DTD is able to discard recurring patterns, e.g. grid lines, wood lines or parking lines. In the fourth image, we see that the method is also robust to some amount of noisy patterns.
While the Sobel filter detects edges reliably, One-Class DTD puts emphasis to edges that are outstanding on a small scale. The scale on which outlierness is detected is parameterized by the patchsize.
\subsection{External Validation}
The following experiment tests the ability of One-Class DTD to produce correct explanations on an artificial problem where we have ground truth information on the input features that cause outlierness.

We build a dataset composed of two horizontally concatenated images of size $28\times 28$. Inliers are constituted by a MNIST digit of the particular digit class on the left, e.g. the class ``0'', and a blank image on the right. A simple one-class SVM with no extension for sequential inputs is trained with a Gaussian kernel and $\sigma=400$ and $\nu=0.01$. After training, the following three cases are considered for explanation: (1) {\em Inlier:} A test image from the training class is presented. (2) {\em Type I outlier:} Structure of the inliers is present (i.e.\ a test example from the training class appears in the left panel) together with some distraction. As distraction, we replace the right panel with a random sample from another random class. (3) {\em Type II outlier:} Structure of the inliers is distracted on both sides; the left panel of type I outliers is replaced by a random sample from another random class.

\begin{figure*}[t!]
\centering
\makebox[\textwidth][c]{\includegraphics[width=1\linewidth]{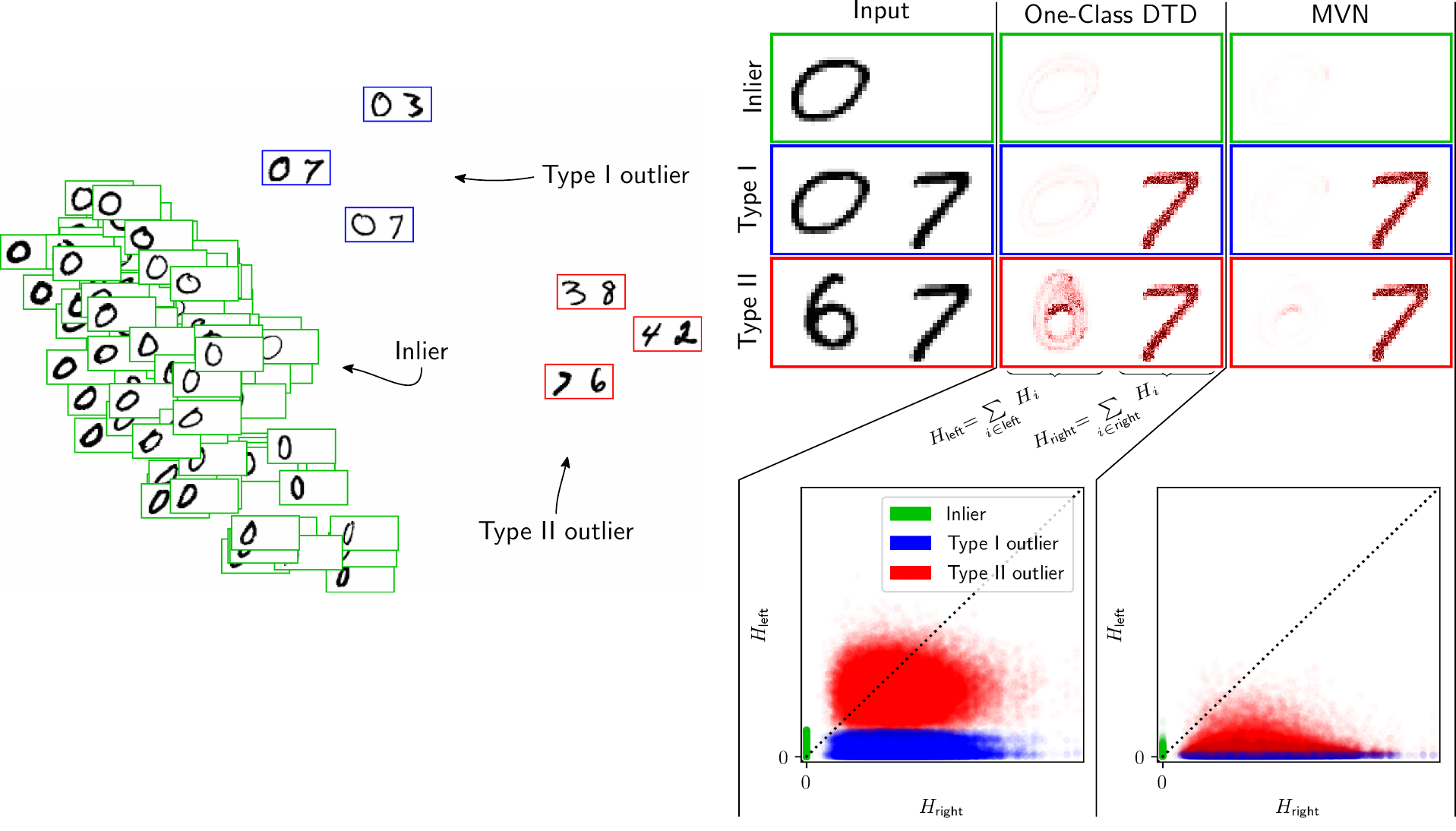}}
\caption{{\em Top:} Random subset of the artificial data set in a 2D PCA embedding for visualization (training and explaining is performed in the original space); only inliers are used for training; {\em Middle:} Explanations from one-class SVM and multivariate normal for one example from every type of pattern; {\em Bottom:} Amount of relevance falling in the left or right side of the image; plot shows results for all ten classes, both models are trained on each class separately; dotted line corresponds to $H_{\textsf{left}}=H_{\textsf{right}}$ share of relevance.}
\label{fig:sel_exp}
\end{figure*}

\Cref{fig:sel_exp} (left) shows some example data for the class of zeros in a 2D PCA embedding.
The ground truth explanation for inliers contains no relevance at all, because the measure of outlierness should detect no evidence for outlierness in these images. 
For type I outliers, the ground truth only contains relevance in the right side of the image. Consider a growing amount of outlierness in the right side only: an explanation of the left side should not be affected by these distractions. 
For type II outliers, relevance should fall in both sides of the image: the left side contains relevance for deviation from the training digit, and the relevance on the right side explains deviation from the blank image. If we consider an input with growing amount of outlierness in the left panel, we see that relevance should also increase in the left panel only and vice versa.

As a baseline, we compare the relevance attribution with the maximum likelihood estimate of a multivariate normal density (MVN) of the training data with no off-diagonal covariances \citep{Murphy:2012:MLP:2380985,bishop1994novelty}. 
The maximum likelihood estimate for the variances is given by $\sigma_i = \frac1n\sum_{j=1}^n(x^{(j)}_i - \mu_i)^2 + \lambda$ with $\boldsymbol{\mu}\in\mathbb{R}^d$ being the mean of training data and a regularization term $\lambda\in\mathbb{R}$. The negative log-likelihood of the MVN, although not a measure of outlierness in the strict sense of Definition \ref{def:outlier}, provides however a natural decomposition on input features as
\begin{align*}
\mathrm{NLL}(\x) = \mathrm{NLL}(\boldsymbol{\mu}) + \sum_{i=1}^d\frac{(x_i - \mu_i)^2}{2\sigma_i} 
\end{align*}
where the first term is a non-decomposable zero-order term and terms of the sum determine the relevance of input features.

\Cref{fig:sel_exp} collects the outcomes of the experiment. In the bottom plots, every sample is represented by one dot. On the $x$-axis, we plot the amount of relevance that falls in the right side of the image, $H_{\textsf{right}}=\sum_{i\in\textsf{right}}H_i$. On the $y$-axis, the relevance of the left image, $H_{\textsf{left}}$, is plotted.
One-class DTD and MVN are both able to explain inliers and type I outliers reliably. They both attribute a small amount of outlierness to the inlier data points, though. The effect of growing outlierness on the right side leading to more relevance in that area can still be observed by looking at blue dots in \Cref{fig:sel_exp} (right).
One-class SVM is better able to explain the outlierness of type II outliers, because it reacts equally strongly to permutations over the input dimensions. 
Instead, the MVN largely ignores outlying patterns in the left panel and thus produces a partial explanation. The incorrect behavior of MVN explanations stems from the fact that the MVN negative log-likelihood is not a true measure of outlierness in the sense of Definition \ref{def:outlier} as it distorts the natural metric of the input space.

\subsection{Internal Validation}\label{sec:pf}
In the following experiments, we consider the output of the one-class SVM as a ground-truth model for outlierness. This allows us to perform validation on datasets for which we do not have a priori knowledge of which features are causing outlierness.

The deep Taylor decomposition method will be compared to a number of other explanation techniques: Sensitivity analysis uses the same trained model but assigns relevance based on the locally evaluated gradient. Other analyses assign relevance based on a simple decomposition of the distance to data, or on the output of some image filter.

For evaluation of explanation quality, we consider the pixel-flipping approach described by \citet{Samek2016} in the context of DNN classifiers. The approach consists of gradually destroying pixels from most to least relevant, and measure how quickly the prediction score decreases.

In the context of outlier detection, however, destroying a pixel does not reduce evidence for outlierness and might even create more of it. Thus, the original pixel-flipping method must be adapted to the specific outlier detection problem. Our approach will consist of performing the flipping procedure not in the pixel-space directly, but in some feature space
\begin{align}
\Psi(\x) = 
 \begin{bmatrix}
 	x_1 - \mathrm{u}^{(1)}_1 & \ldots & x_1 - \mathrm{u}^{(m)}_1 \\
 	\vdots & \ddots & \vdots\\
 	x_d - \mathrm{u}^{(1)}_d & \ldots & x_d - \mathrm{u}^{(m)}_d
 \end{bmatrix}.
\label{eq:fspf}
\end{align}
containing all component-wise differences to support vectors. The one-class SVM can be rewritten in terms of elements of this feature space as $g(\Psi) = \sum_j \alpha_j k(\|\Psi_{:,j}\|)$, and similarly the outlier function can be written as $o(\Psi)$.

Our modified procedure reduces the dimensionality of the data one dimension at the time. Once dimensionality 0 is reached, the pattern is necessarily an inlier, because no deviance from the support vectors exists anymore. This method makes removal of outlierness computational feasible. The ordering of variables is inferred from the relevance scores assigned to each input dimension (cf.\ \ref{appendix:pixelflipping} for pseudo-code). Also note that we seek to provide a {\em global} explanation of the outlierness of pattern $\x$. Except for trivial cases (with only one support vector) no single pattern in $\mathcal{X}$ can represent a minimizer of all detectors that the model is composed of.

We train a one-class SVM on the CIFAR-10 data set. The data set consists of 50000 images in the space $\mathbb{R}^{32\times 32\times 3}$ with values ranging from 0 to 255. The images are divided into 625 patches of adjacent pixels each, where a patch is of dimensionality $7\times 7\times 3$. This leads to more than 31 million training vectors $\x_i \in \mathbb{N}^{147}$. For training speedup, we randomly select 30,000 patches from the data set and train a one-class SVM on these patches. The outlier scores are summed over all patches of an image, as described in \Cref{sec:seq}, to get a measure of outlierness for the whole image.
\begin{figure*}
\makebox[\textwidth][c]{\includegraphics[width=1\textwidth]{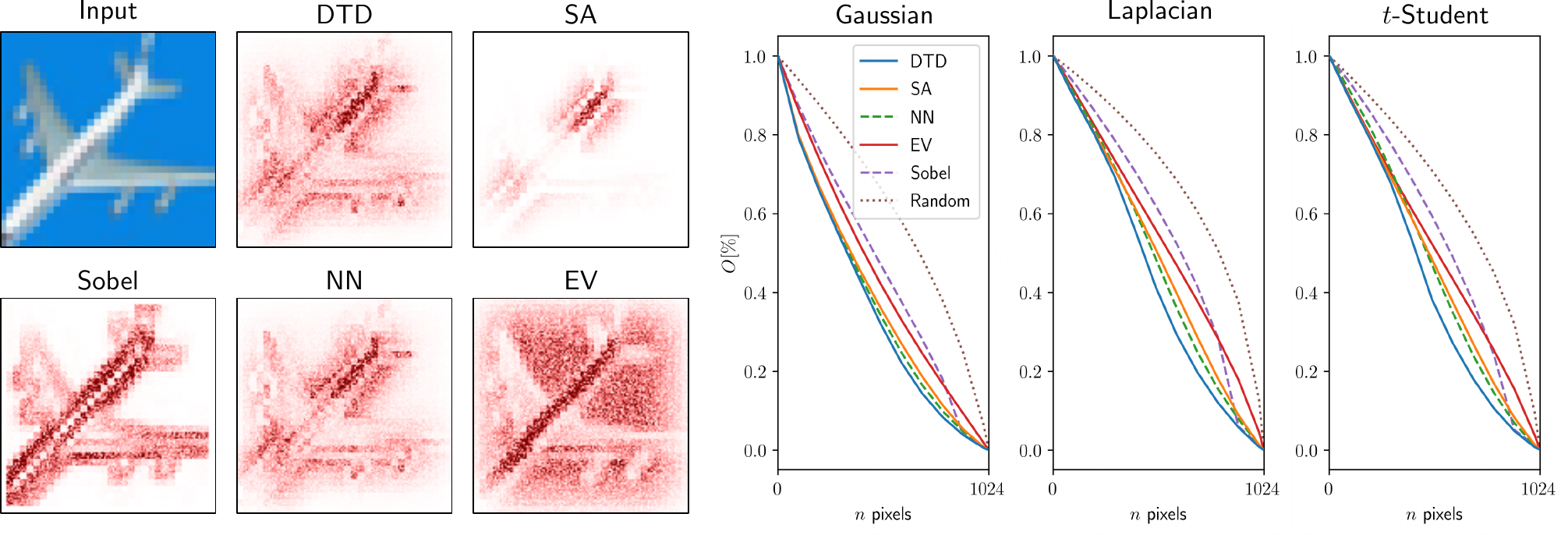}}
\caption{Pixel flipping experiment; {\em Left:} Example from the CIFAR-10 class ``airplane'' shown next to the explanation and several baselines; here shown for the Gaussian kernel; {\em Right:} Pixel flipping experiment for several kernels; $t$-Student is with $q=1$.}\label{fig:pf}
\end{figure*}
\Cref{fig:pf} shows an example image and heatmaps from all attribution methods.

We consider as baselines for explanation sensitivity analysis (SA) as defined in \Cref{section:background}, the squared difference to the nearest neighbor (NN), the squared difference to the expected pixel value (EV) and the Sobel filter. The squared difference to the nearest neighbor support vector (NN),
\[
\mathbf{R}^{\mathrm{NN}} = (\x - \sv_{\mathrm{NN}})^2
\]
with $\sv_{\mathrm{NN}} = \argmin_{\sv_j}\|\x - \sv_j\|$ is similar to DTD but performs a min-take-all redistribution instead of min-take-most. This yields discontinuities in the explanation for perturbations of the inputs. This issue is reduced in the sequential model due to the overlap of patches, however.
We also add the squared difference to the expected pixel value (EV) to the baselines.
EV is inferred from the support vectors by
\[
\mathbf{R}^{\mathrm{EV}} = \left(\x - \bar{\sv}\right)^2
\]
where $\bar{\sv} = \sum_{j=1}^m\alpha_j\sv_j$.
Finally, a random pixel ordering is considered as a completely uninformed baseline method.
\Cref{fig:pf} shows the results of the pixel flipping experiment for all methods and several kernels. Deep Taylor decomposition is indeed superior for all considered kernels.
Sensitivity analysis can be interpreted as the explanation of local variation of the detection function in the vicinity of the pattern in question. We can see that the local gradient is not as well suited for explanation as DTD. In particular, sensitivity is unable to detect truly relevant pixels that cause the outlier score to be large. 
Instead, it assigns the most relevance to pixels to which the model is sensitive {\em locally} (cf.\ \citet{Samek2016}). As mentioned before, nearest neighbor support vectors provide an explanation that is discontinuous to perturbations of the inputs. Explanations from the NN procedure are more complete compared to SA. As we see in the right plots of \Cref{fig:pf}, the pixel ordering is still diverging from the explanation that is produced by DTD.
The EV baseline corresponds roughly to a squared difference to the data mean, and is even more global than DTD. We see that its performance in the pixel flipping experiment is still better than Sobel and Random flipping.
Remaining baselines (EV, Sobel, Random) fail to produce a competitive explanation.

Results for more kernels can be found in \ref{sec:pf_all}.
\subsection{Intrusion detection}
One-class SVM has been applied to network intrusion detection and malware detection \cite{DBLP:journals/jair/GoernitzKRB13,wressnegger2013close,rieck2009machine,wang2004anomaly,DBLP:journals/tse/Denning87}. Having interpretable model outputs can help to identify the intent or the method of an attack. We take up this idea in a simpler setting where no domain knowledge is necessary and where it is arguably possible to detect outlierness on a symbolic level, that can be compared to an attack. In particular, we train a one-class SVM on the personal attacks corpus from the Detox data set \citep{DBLP:journals/corr/WulczynTD16}. 
In this dataset, documents are labeled by up to ten annotators as either 0 (neutral) or 1 (personal attack). 

A dictionary is constructed from stemmed terms that appear in at least five documents and binary features are extracted as a vectorial representation of documents. No stop words are removed and no document frequencies are used for feature extraction. The model is trained on samples with label mean 0 with a Gaussian kernel and $\nu = 0.3$. Parameter $\sigma$ is set to 10, which is a soft assumption of an expected difference in 10 terms for similar documents. 

\begin{figure}
\centering 
\fbox{\includegraphics[width=.97\linewidth]{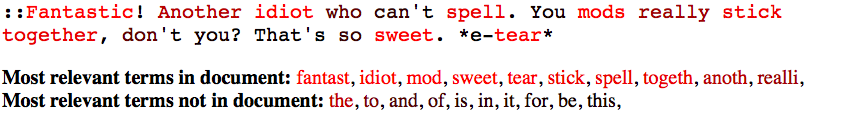}}
\fbox{\includegraphics[width=.97\linewidth]{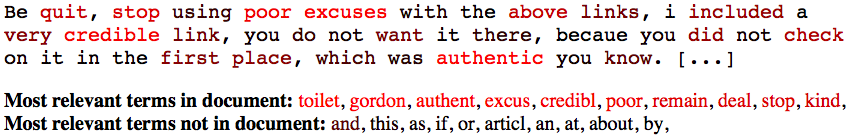}}
\caption{Relevance assignment for two sample message from the Detox data set; red color indicates relevance scores. Below each document, most frequent terms in the document and most frequent terms that are not in the document are listed.}\label{fig:messages}
\end{figure}

Interpretable outputs are produced in terms of term relevance scores \cite{arras-plos17,DBLP:journals/corr/HornAMMS17}.
\Cref{fig:messages} shows the explanation for two example documents. As one would expect, common terms have no or low relevance in the document and terms that would not be expected in a neutral message receive more relevance. Due to the RBF property, relevance will also be assigned to terms that do not appear in a document. These terms can be interpreted as being benign and expected to appear in a typical example. This quantity can be of interest in text analysis and could not be derived from, e.g., a linear model.
Note the ironic use of the word {\em fantastic}. The term receives most relevance, simply because it is not used frequently in neutral messages. The interpretation of the term being detected due to the ironic use can not be justified for such a symbolic model. The property to assign high relevance to rare events is still given. Rare events, here, is the presence of terms which appear rarely or missing terms that usually appear. Outlierness continues to grow if more rare events appear \cite{DBLP:journals/corr/abs-1708-08296}.

\section{Conclusion}
\label{section:conclusion}
{
In this paper, we have addressed the problem of anomaly explanation. Technically, we have proposed a deep Taylor decomposition of the one-class SVM. It is applicable to a number of commonly used kernels, and produces explanations in terms of support vectors or input variables. Our empirical analysis has demonstrated that the proposed method is able to reliably explain a wide range of outliers, and that these explanations are more robust than those obtained by sensitivity analysis or nearest neighbor.

A crucial aspect of our explanation method is that it required us to elicit a natural neural network architecture for the problem at hand. Achieving this in the context of the one-class SVM model has highlighted the asymmetry between the problem of inlier and outlier detection, where the first one can be modeled as a max-pooling over similarities, and where the latter is better modeled as a min-pooling over distances. The novel insight on the structure of the outlier detection problem might inspire the design of deeper and more structured outlier detection models.
}

\section*{Acknowledgments}

This work was supported by the Brain Korea 21 Plus Program through the National Research Foundation of Korea; the Institute for Information \& Communications Technology Promotion (IITP) grant funded by the Korea government (MSIT) [No. 2017-0-01779]; the Deutsche Forschungsgemeinschaft (DFG) [grant MU 987/17-1]; and the German Ministry for Education and Research as Berlin Big Data Center (BBDC) [01IS14013A]. This publication only reflects the authors views. Funding agencies are not liable for any use that may be made of the information contained herein.
We are grateful to Guido Schwenk for the valuable discussion.

{
\small
\bibliographystyle{abbrvnat}
\bibliography{oneclass}
}


\appendix
\section{Pseudo codes}
\label{appendix:preimages}
In this section, we list pseudo codes for the proposed algorithms.

\subsection{Support Vector Relevance}
Support vector relevance is the decomposition of inliers and the higher-layer relevance for outliers. All decompositions can be calculated in terms of quantities that are already computed in the forward path.
\begin{algorithm}[H]
\begin{algorithmic}[0]
\Instart
\State $\texttt{A} = (\alpha_j)_j$ \Comment{Weight vector $\boldsymbol{\alpha}$}
\State $\texttt{K} = (k(\sv_j, \x))_j$ \Comment{Vector of kernel evaluations at $\x$}

\Inend
\Outstart
\State $(R_j)_j$ \Comment{Support vector relevances}
\Outend
\Procedure{}{}
\State $(R_j)_j \gets (\alpha_j k(\|\x-\sv_j\|))_j$ for all $j=1,\ldots,m$
\EndProcedure
\end{algorithmic}
\caption{Inlier explanation}
\end{algorithm}
\subsection{Input relevance for $t$-Student kernels}
For the $t$-Student kernel, we identify decomposable upper-layer relevance as
\[
\Delta_j = R_j - D_j^{+} = \frac{\alpha_jk(\sv_j,\x)}{\sum_{j'=1}^m\alpha_{j'}k(\sv_{j'},\x)}\cdot\frac{\|\x-\sv_j\|^q}{a+\|\x-\sv_j\|^q}\cdot o
\]
\begin{algorithm}[H]
\begin{algorithmic}[0]
\Instart
\State $\x$ \Comment{Input vector $\x$}
\State $(\sv_j)_j$ \Comment{$m\times d$ matrix of support vectors}
\State $(\alpha_j)_j$ \Comment{Weight vector $\boldsymbol{\alpha}$}
\Inend
\Outstart
\State $\mathbf{R}$ \Comment{Input relevance vector}
\Outend
\Procedure{}{}
\State $\texttt{d} \gets (\|\x-\sv_j\|^q)_j$
\State $\texttt{h} \gets (\alpha_j^{-1}\cdot(a+\texttt{d}_j))_j$
\State $\texttt{o} \gets m\slash(\mathbf{1}^\top \texttt{h}^{-1})$
\State $\Delta \gets \frac{\texttt h^{-1}}{\mathbf{1}^{\top}\texttt h^{-1}}\odot \frac{\texttt d}{a+\texttt d}\odot\texttt{o}$
\State $\texttt{S}_{ji} \gets \left(x_i-\mathrm{u}^{(j)}_i\right)^2$ for all $i=1,\ldots,d$, $j=1,\ldots,m$
\State $\mathbf{R} \gets \frac{\texttt S^{\top}}{\mathbf{1}^{\top}\texttt S}\Delta$
\EndProcedure
\end{algorithmic}
\caption{Outlier explanation for $t$-Student kernels}
\end{algorithm}
Here, $\odot$ denotes element wise products, $\oslash$ the element wise division. The matrix $\texttt{V}$ and vectors  $\texttt{d}$ and $\texttt{k}$ can be precomputed for better runtime performance.

\subsection{Input relevance for exponential kernels}
For the exponential kernels, the decomposable upper-layer relevance can be written as
\[
\Delta_j = R_j - D_j^+ = \frac{\alpha_jk(\sv_j,\x)}{\sum_{j'=1}^m\alpha_{j'}k(\sv_{j'},\x)}\cdot\min\left(o, \frac{\|\x-\sv_j\|^q}{q\sigma^q}\right)
\]
This allows the following fast algorithm for input relevance.
\begin{algorithm}[H]
\begin{algorithmic}[0]
\Instart
\State $\x$ \Comment{Input vector $\x$}
\State $(\sv_j)_j$ \Comment{$m\times d$ matrix of support vectors}
\State $(\alpha_j)_j$ \Comment{Weight vector $\boldsymbol{\alpha}$}
\Inend
\Outstart
\State $\mathbf{R}$ \Comment{Input relevance vector}
\Outend
\Procedure{}{}
\State $\texttt{d} \gets (\|\x-\sv_j\|^{q}/q\sigma^q)_j$
\State $\texttt{h} \gets (-\texttt{log}(\alpha_j)+\texttt{d}_j)_j$
\State $\texttt{o} \gets -\texttt{LSE}(-\texttt{h})$
\State $\Delta \gets \frac{\texttt{exp}(-\texttt{h})}{\mathbf{1}^\top\texttt{exp}(-\texttt{h})}\odot\texttt{min}\left(\texttt o, \texttt{d}\right)$
\State $\texttt{S}_{ji} \gets \left(x_i-\mathrm{u}^{(j)}_i\right)^2$ for all $i=1,\ldots,d$, $j=1,\ldots,m$
\State $\mathbf{R} \gets \frac{\texttt S^{\top}}{\mathbf{1}^{\top}\texttt S}\Delta$
\EndProcedure
\end{algorithmic}
\caption{Outlier explanation for exponential kernels}\label{alg:expout}
\end{algorithm}
Here $\odot$ denotes element wise multiplication.

\subsection{Pixel flipping procedure}\label{appendix:pixelflipping}
We show here a pseudo-code of the pixel flipping experiment that we perform in \Cref{sec:pf}.
\begin{algorithm}[H]
\begin{algorithmic}
\Instart
\State $\Psi(\x)$ \Comment{Effective inputs}
\State $\mathbf{R}$ \Comment{Heatmap}
\Inend
\Outstart
\State \texttt{pfcurve} \Comment{Declining outlier score}
\Outend
\Procedure{}{}
\State {\tt pfcurve} $\gets$ {\tt [\,]}
\For{{\tt i} \textbf{in} {\tt argsort}($-\mathbf{R}$)}
\State $\Psi_{\texttt{i},\texttt{:}} \gets \boldsymbol{0}$
\State {\tt pfcurve.append}($o(\Psi)$)
\EndFor
\State \Return {\tt pfcurve}
\EndProcedure
\end{algorithmic}
\caption{Pixel flipping procedure}
\end{algorithm}
\noindent where $\mathbf{R}$ is the heatmap to evaluate and ``{\tt pfcurve}'' is the result of the analysis.

\section{Proofs}
In this section we prove asymptotic convergence for our proposed measures of outlierness, equality of proposed neural networks with kernelized one-class SVM and a unified formulation of support vector relevance for measures of outlierness.
\subsection{Proofs for outlierness measures of \Cref{sec:quantifying}}\label{sec:out_conform}
In the following, we show that outlierness measures constructed from the one-class SVM with the $t$-Student and exponential kernels satisfy \Cref{def:outlier}. For both proofs we use $\sum_{j=1}^m\alpha_j = 1$.
First, we consider the $t$-Student kernel. Condition 1 follows from the positivity of $g(\x)$. Condition 2 is proven below.
\begin{align*}
	\frac{o(t\x)}{\|t\x\|^q} &\overset{\hphantom{t\rightarrow\infty}}{=} \frac{m}{\|t\x\|^q\cdot\sum_{j=1}^m\alpha_j(a+\|t\x-\sv_j\|^q)^{-1}}\\
	&\overset{t\rightarrow\infty}{=} \frac{m}{t^q\cdot\sum_{j=1}^m\alpha_jt^{-q}} = m
\end{align*}
Next, we show the convergence for exponential kernels. Condition 1 follows from $g(\x)$ being upper bounded by 1 for the exponential kernels. The proof of the second condition follows below.
\begin{align*}
	\frac{o(t\x)}{\|t\x\|^q} &\overset{\hphantom{t\rightarrow\infty}}{=} \frac{-\log\left(\sum_{j=1}^m\alpha_j\exp\left(-\frac{\|t\x-\sv_j\|^q}{q\sigma^q}\right)\right)}{\|t\x\|^q}\\
	&\overset{t\rightarrow\infty}{=}\frac{-\log\left(\sum_{j=1}^m\alpha_j\exp\left(-t^q\right)\right)}{t^q}=1
\end{align*}

\subsection{Equivalence of the one-class SVM with the neural network representation}
\label{sec:nnrepproof}
We show that the neural network from \Cref{section:deeptaylor-student} implements the measure of outlierness for the $t$-Student kernel that is proposed in \Cref{sec:quantifying}.

Let $k$ be the $t$-Student kernel and $h_j = \frac{1}{\alpha_j} (a+\|\x - \sv_j\|^q)$, then
\begin{align*}
o &= \mathrm{H}((h_j)_j)= \frac{m}{\sum_{j=1}^m h_j^{-1}}\\
&= \frac{m}{\sum_{j=1}^m \alpha_j \frac{1}{a + \|\x - \sv_j\|^q}} = \frac{m}{g(\x)}.
\end{align*}
Next, we show the equivalence for exponential kernels. Let therefore $k$ be the exponential kernel and $h_j = -\log(\alpha_j) + \frac{1}{q}\left(\frac{\|\x - \sv_j\|}{\sigma}\right)^q$. Then
\begin{align*}
o &= -\mathrm{LSE}((-h_{j})_{j}) = -\log\left(\sum_{j=1}^m \exp(-h_j)\right) \\
&= -\log\left(\sum_{j=1}^m \alpha_j e^{-\frac{\|\x - \sv_j\|^q}{q\sigma^q}}\right) = -\log(g(\x))
\end{align*}
\subsection{Decompositions of $o$ are conservative}\label{sec:linproof}
We prove $\sum_{j=1}^mR_j = o$. First, for $t$-Student kernels:
\begin{align*}
\sum_{j=1}^m R_j &=\sum_{j=1}^m h_j\cdot \frac{m\cdot\left(\frac{1}{h_j}\right)^2}{\left(\sum_{j'=1}^m \frac{1}{h_{j'}}\right)^2}\\
&=\sum_{j=1}^m \frac{1}{h_j} \cdot \frac{m}{\left(\sum_{j'=1}^m\frac{1}{h_{j'}}\right)^2}\\
&= \mathrm{H}((h_j)_j) = o.
\end{align*}
Next, we show the conservation for exponential kernels.
\begin{align*}
\sum_{j=1}^m R_j &=	\sum_{j=1}^m\frac{\exp(-h_j)}{\sum_{j'=1}^m \exp(-h_{j'})} \cdot (-\mathrm{LSE}((-h_{j'})_{j'})\\
&= -\mathrm{LSE}((-h_{j})_{j}) = o
\end{align*}
As a consequence, all higher order terms in the Taylor expansion sum to zero.
\subsection{Constancy of $c_j$, $p_j$ and $\varepsilon_j$}\label{sec:constant}
First, we show the invariance of $c_j$ with respect to scaling. Let $(t\cdot h_j)_j$ for some $t$ be a scaling of $h_j$.
\begin{align*}
c_j(t) = \mathrm{H}\bigg(\bigg(\frac{t\cdot h_{j'}}{t\cdot h_j}\bigg)_{j'}\bigg)^2 = \mathrm{H}\bigg(\bigg(\frac{h_{j'}}{h_j}\bigg)_{j'}\bigg)^2 = c_j
\end{align*}
Next, we prove that $p_j$ and $\varepsilon_j$ from \Cref{section:deeptaylor-exponential} are constant for any constant increment. Let therefore $(h_j + t)_j$ for some $t$ be the incremented $h_j$.
\begin{align*}
p_j(t) &= \frac{\exp(-(h_j+t))}{\sum_{j'=1}^m\exp(-(h_{j'}+t))} \\
&= \frac{\exp(-t)\cdot\exp(-h_j)}{\exp(-t)\cdot\sum_{j'=1}^m\exp(-h_{j'})}\\
&= \frac{\exp(-h_j)}{\sum_{j'=1}^m\exp(-h_{j'})} = p_j
\end{align*}
\begin{align*}
\varepsilon_j(t) &= -\mathrm{LSE}(-((h_{j'}+t)-(h_j+t))_{j'})\\
&= -\mathrm{LSE}(-(h_{j'}-h_j))	= \varepsilon_j
\end{align*}
It follows that $c_j, p_j$ and $\varepsilon_j$ stay constant for the transformations that we discuss in \Cref{section:deeptaylor-input}.
\subsection{Decomposition of $R_j$}\label{sec:dec_rj}
In this section, we derive the decomposition of 
\begin{align}
R_j = C_j\cdot\|\x-\sv_j\|^q + D_j\label{eq:rjofx}
\end{align} in terms of input variables $\x$ in order to elaborate on some critical steps.
For that, we show that integrated gradients \cite{DBLP:journals/corr/SundararajanTY17} of $R_j$ has a closed analytic solution and does not need to be calculated numerically.
The integrated gradients are formally defined as
\begin{align}
	\mathbf{R}_j &= (\x-\xroot_j)\odot\int_{0}^1 \frac{\partial R_j}{\partial \x}\Big|_{\xroot_j + t(\x-\xroot_j)}dt \label{eq:intgraddef}
\end{align}
which, as a consequence of the gradient theorem, is a conservative decomposition of $R_j(\x)-R_j(\xroot_j)$. If $\xroot_j$ is also a root of $R_j$, then integrated gradients can serve as an explanation of $R_j(\x)$. The coefficient $C_j$ is always positive.
We need to consider separately the case where $D_j>0$, $D_j=0$ and $D_j<0$.
\begin{enumerate}
\item When $D_j > 0$, there is no root point, but $R_j$ still admits a minimum at $\sv_j$. Performing integrated gradients at this minimum is possible but the decomposition will not be conservative. 
\item When $D_j = 0$, the relevance function has a root point at $\x = \sv_j$.
\item When $D_j < 0$, there is always a root point of the relevance function. The nearest root point from $\x$ is the one on the segment between the data point $\x$ and the support vector $\sv_j$.
\end{enumerate}
All these cases can be regrouped in the same redistribution formulas from support vector relevances to the heatmap $\mathbf{R}$ in input domain.
We prove the decomposition for $D_j < 0$ and subsequently generalize to the other two cases. Assume $D_j < 0$ and $C_j > 0$.

\begin{enumerate}
\item The gradient of $R_j=C_j\|\x-\sv_j\|^q + D_j$ w.r.t. $\x$ can be written as 
\begin{align*}
\frac{\partial R_j}{\partial \x} =  q\cdot\frac{\boldsymbol{z}_j}{\|\boldsymbol{z}_j\|}\cdot C_j\|\boldsymbol{z}_j\|^{q-1}
\end{align*}
with $\boldsymbol{z}_j = \x-\sv_j$.
\item The nearest root of $R_j$ lies on the segment $[\sv_j,\x]$ (and exists by assumption). Integrating that gradient between $r\boldsymbol{z}_j$ and $\boldsymbol{z}_j$ for some $0<r<1$ (i.e. on the segment) yields
\begin{align}
&\int_{r\boldsymbol{z}_j}^{\boldsymbol{z}_j} q\cdot\frac{\boldsymbol{z}_j}{\|\boldsymbol{z}_j\|}\cdot C_j\|\boldsymbol{z}_j\|^{q-1} d\boldsymbol{z}_j \notag\\
&= \int_{0}^{1}\frac{\boldsymbol{z}_j}{\|\boldsymbol{z}_j\|}\cdot C_j\|\boldsymbol{z}_j\|^{q-1}\cdot q\cdot(r+t(1-r))^{q-1} dt \notag\\
&= \frac{\boldsymbol{z}_j}{\|\boldsymbol{z}_j\|}\cdot C_j\|\boldsymbol{z}_j\|^{q-1}\cdot\int_{0}^{1}q\cdot(r+t(1-r))^{q-1}dt\notag\\
&= \frac{\boldsymbol{z}_j}{\|\boldsymbol{z}_j\|}\cdot C_j\|\boldsymbol{z}_j\|^{q-1}\cdot\frac{1-r^q}{1-r} \label{eq:intgrad}
\end{align}
\item Decomposition of $R_j$ by integrated gradients \cite{DBLP:journals/corr/SundararajanTY17} is given by combining \Cref{eq:intgraddef,eq:intgrad} and parametrization $\xroot_j = \sv_j+r(\x-\sv_j)$:
\begin{align*}
\mathbf{R}_j &= (\x-\xroot_j)\odot\int_{0}^1 \frac{\partial R_j}{\partial \x}\Big|_{\xroot_j + t(\x-\xroot_j)}dt \\
&= (\boldsymbol{z}_j - r\boldsymbol{z}_j)\odot \frac{\boldsymbol{z}_j}{\|\boldsymbol{z}_j\|}\cdot C_j\|\boldsymbol{z}_j\|^{q-1}\cdot\frac{1-r^q}{1-r}\\
&= \frac{\boldsymbol{z}_j^2}{\|\boldsymbol{z}_j\|^2}\cdot C_j\|\boldsymbol{z}_j\|^{q}\cdot(1-r^q)
\end{align*}
\item Injecting for $r$ the root of \Cref{eq:rjofx}, $\widetilde{\boldsymbol{z}}_j=r\boldsymbol{z}_j$, with
\begin{align*}
r^q = \frac1{\|\boldsymbol{z}_j\|^q}\cdot\left(-\frac{D_j}{C_j}\right)
\end{align*}
gives the decomposition
\begin{align*}
\mathbf{R}_j = \frac{\boldsymbol{z}_j^2}{\|\boldsymbol{z}_j\|^2}R_j = \left[\frac{\x-\sv_j}{\|\x-\sv_j\|}\right]^{2}\cdot R_j.
\end{align*}
\end{enumerate}
Of course, the root $\widetilde{\boldsymbol{z}}_j=r\boldsymbol{z}_j$ only exists if $D_j$ and $C_j$ have opposite sign. When $D_j$ is positive, $R_j$ has no root and the global minimum at $\xroot_j=\sv_j$ serves as the reference for decomposition. This decomposition will not be conservative, because the term $R_j(\xroot_j)=R_j(\sv_j)=D_j$ will be strictly positive. Collecting all cases, the decomposition of $R_j$ on the input will be given by
\begin{align*}
\mathbf{R}_j = \left[\frac{\x-\sv_j}{\|\x-\sv_j\|}\right]^{2}\cdot(R_j-D_j^+)
\end{align*}
where $D_j^+ = \max(0, D_j)$.
Since $\mathbf{R} = \sum_j \mathbf{R}_{j}$ (cf. \cite{10.1371/journal.pone.0130140} for the details), the full input relevance can be written as
\begin{align*}
\mathbf{R} = \sum_{j=1}^m \left[ \frac{\x-\sv_j}{\|\x-\sv_j\|} \right]^2\cdot(R_j - D_j^+).
\end{align*}

\section{Quantitative results for several more kernels}\label{sec:pf_all}
\begin{figure}[h]
\includegraphics[width=1\linewidth]{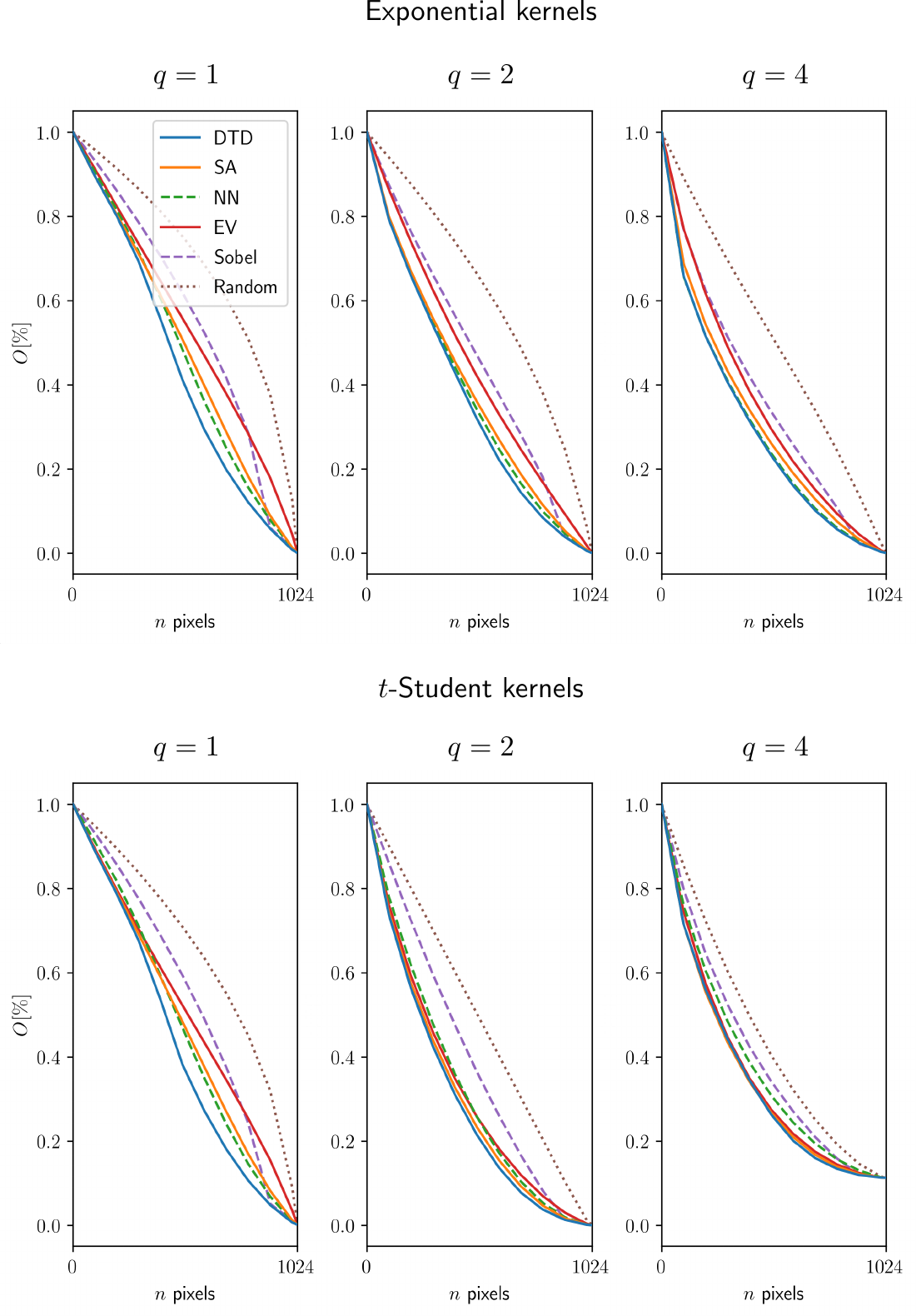}
\caption{Results of the pixel flipping experiment for several kernels and parameters.}
\label{fig:pf_all}
\end{figure}

In this section we collect the results of selectivity experiments from \Cref{sec:pf} for several more kernels, in particular for both, exponential and $t$-Student kernels, we show results for $q=1,2,4$. Note that for $q=4$ or larger, the kernel matrix tends to be singular. We observe a clear tendency of NN and DTD converging to the same performance for the exponential kernels. Overall, DTD performs most reliably over all cases that we consider here.
\Cref{fig:pf_all} shows the results on the CIFAR-10 dataset as described in the main paper. 
\end{document}